\documentclass[letterpaper]{article}
\usepackage{flairs}
\usepackage{times}
\usepackage{helvet}
\usepackage{courier}
\usepackage{graphicx}
\usepackage{url}

\begin{document}
\title{Propasafe-Hybrid: A Text-Based Hybrid Propaganda Detection Tool}
\author{Thomas Kimmeth, Avijit Roy, Vivek Sharma\\
John Jay College of Criminal Justice, CUNY\\
New York, NY\\
aroy@jjay.cuny.edu\\
}

\maketitle

\begin{abstract}
Propagandistic content increasingly circulates through online news and social media, where readers often encounter it with limited scrutiny, highlighting the need for reliable and fine‑grained detection. This paper introduces Propasafe‑Hybrid, a sentence‑level system that integrates a fine‑tuned transformer classifier with LLM‑based technique classification to identify, label, and explain specific propaganda strategies. The pipeline generates actionable outputs, including highlighted sentences, technique assignments, and concise rationales, so users can immediately understand why a sentence was flagged and how each label was determined. To control inference cost, Propasafe‑Hybrid employs a cost‑aware pre‑filtering stage that forwards only high‑likelihood sentences to LLMs, reducing token usage while preserving the underlying decision logic. Together, these design choices enhance the explainability, efficiency, and practical usability of sentence‑level propaganda detection in real‑world news environments.
\end{abstract}

\section{Introduction}
\label{sec:intro}

The contemporary information ecosystem characterized by frictionless publishing, algorithmic amplification, and coordinated online networks, has intensified the reach and impact of propaganda in news and social media. Rather than focusing solely on veracity, computational propaganda campaigns typically seek to shape beliefs and behaviors by leveraging persuasive techniques, anonymity, micro‑targeting, and recommendation dynamics to influence salient public issues \cite{martino2020survey}. At the textual level, research has moved from coarse document‑level labels toward fine‑grained, technique‑level analysis (e.g., loaded language, name‑calling, flag‑waving), enabling span‑level explanations that make manipulation visible where it occurs \cite{da2019fine,da2020semeval}. This reframing was consolidated by SemEval‑2020 Task 11, which standardized span identification and technique classification and demonstrated the effectiveness of Transformer‑based approaches under shared, span‑aware evaluation protocols \cite{da2020semeval}.

Despite these advances, a practical gap remains between research prototypes and tools that ordinary readers can use with appropriate privacy guarantees. Early public systems such as Proppy showed that stylistic features can score propagandistic style at the article level, but offered limited sentence‑level transparency and have been inconsistently available over time \cite{barron2019proppy}. Newer interactive platforms like PRTA~\cite{da2020prta} for span‑level exploration, and browser extensions like Apollolytics~\cite{zavolokina2025biased} that highlight techniques and provide explanations, brings detection to the point of consumption, but typically rely solely on server‑side large language models (LLMs), raising concerns about cost, latency, and data transfer. 
Human‑computer interaction work further indicates that explainable nudges can foster more analytical (“System‑2”) reading, suggesting that usable explanations, not just labels, are central to impact \cite{zavolokina2024think}. 

These tensions motivate solutions that (i) deliver on‑page, real‑time feedback where readers encounter content, and (ii) balance explainability and sustainability by reducing token consumption on third‑party APIs. Privacy‑first prototypes such as Propasafe\footnote{\url{https://github.com/wewakesharma/propasafe}} show that locally fine‑tuned BERT models can provide sentence‑level cues offline, complementing cloud‑backed, explainable systems like Apollolytics. In short, the motivation is to equip readers with transparent and optimized token usage of APIs that exposes persuasive techniques at the granularity where they operate.

Building on this landscape, our work extends the publicly available Propasafe tool by preserving its offline, privacy‑conscious architecture and adding model‑supported explanations of detected spans, thereby combining the advantages of local inference with the interpretability benefits demonstrated, while reducing the number of tokens used in recent HCI‑driven systems.

\section{Background}
\label{sec:background}

Detecting propaganda in news and online discourse has increasingly become a central task in computational social science and NLP. Early research highlighted the limitations of document‑level labels, which often failed to reveal where and how manipulation occurs within text. This motivated the creation of fine‑grained resources such as the Propaganda Techniques Corpus (PTC)\cite{da2019fine} and SemEval 2020 Task 11 competitions\cite{da2020semeval}. These efforts demonstrated that models such as BERT, when fine‑tuned on PTC, can identify propagandistic cues at the sentence or span level, offering a more interpretable and actionable view of manipulation in text.

\begin{figure}[!tbp]
    \centering
    \includegraphics[width=0.9\columnwidth]{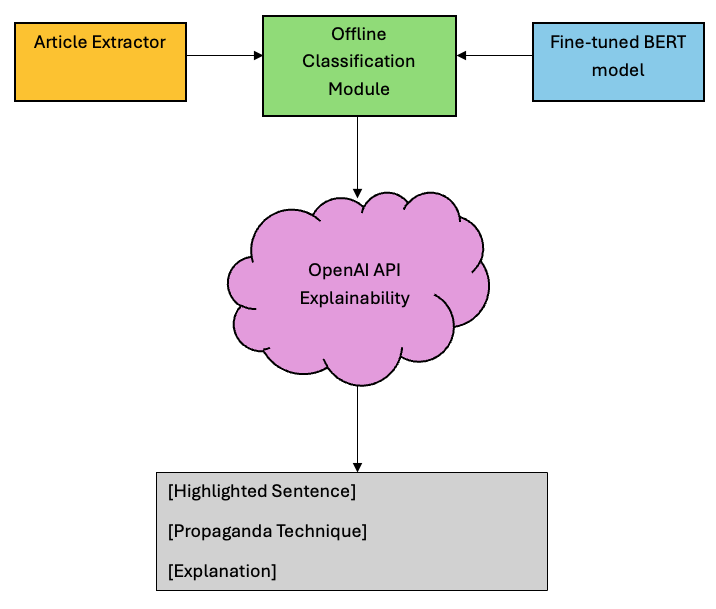}
    \alt{Block diagram showing the Propasafe-Hybrid architecture. Three inherited components from the original Propasafe system are shown at the top: Article Extractor, Offline Classification Module, and Fine-tuned BERT model. These feed into a newly introduced OpenAI API Explainability module. The final output consists of three elements: Highlighted Sentence, Propaganda Technique, and Explanation.}
    \caption{Architecture of Propasafe-Hybrid showing the inherited components from the original Propasafe system and the newly introduced Explainability Module.}
    \label{fig:architecture_img}
\end{figure}

Alongside academic progress, a growing ecosystem of browser‑based propaganda detection tools has emerged. These tools aim to bring real‑time assistance directly to readers as they navigate news articles and social media platforms. Some representative examples include

\begin{itemize}
    \item{Proppy~\cite{barron2019proppy} is a real‑time propaganda scoring for online news at document level. It used stylistic features like character n‑grams, readability, and keywords, trained on known propaganda sources. The tool organized articles into events and ranked them by propagandistic likelihood. The original public site has been intermittently unavailable, but papers is publicly available.}
    \item{Propasafe~\cite{sharma2025propasafe} is a privacy‑focused, offline browser extension using fine‑tuned BERT on the PTC data for sentence‑level classification. It does not use online inference and lacks explainability; the result is binary (propaganda or non-propaganda).}
    \item{Apollolytics~\cite{zavolokina2025biased} is a cross‑browser extension that flags techniques and contextualizes claims with sources. It has a variety of options to choose from for the models. The tool highlights and explains the techniques used in the sentences.}
    \item{Conscious Chrome Extension provides a tweet-level analysis for propaganda, reliability, and perspective via OpenAI API. It sends text to GPT3.5/4 and aggregates supporting news via APIs.}
\end{itemize}

\section{Architecture}
\label{sec:architecture}

Our tool builds upon the modular framework originally introduced in Propasafe~\cite{sharma2025propasafe}, which provides a lightweight and extensible architecture for offline inference of propagandistic content. Its modular design allowed us to reuse existing components, particularly the article extraction and local BERT-based classification pipeline, while incorporating new functionality for explainability.

The Article Extractor fetches raw HTML via \texttt{requests} and applies a four-stage cascade: DOM pre-cleaning with \texttt{lxml} (removing modals, ads, navbars, and figure captions, then isolating the \texttt{<article>} subtree when present), followed by text extraction attempted in priority order using \texttt{trafilatura}, \texttt{readability-lxml}, and \texttt{newspaper3k}. The resulting text is normalized with regex-based sentence boundary fixes before tokenization via NLTK's Punkt tokenizer, with a fragment-stitching pass to correct over-splits on courtesy titles. The final sentence list (capped at 400 sentences) is passed directly to the BERT classifier.
In the Propasafe framework, the offline inference module processes each sentence using a fine‑tuned BERT model and outputs a binary decision indicating whether the sentence contains propaganda. We leverage this capability as a filtering layer, where only sentences that the offline model marks as propagandistic are forwarded to the next stage. As noted in the original Propasafe paper, the overall quality of detection depends heavily on the specific model loaded into the offline module; the authors report a default BERT configuration achieving an F1 score of approximately 0.71~\cite{sharma2025analysis} on the PTC dataset.

To expand this baseline, we introduce an explainability module powered by LLM APIs. After the offline classifier identifies potential propaganda sentences, those sentences are passed to a prompt‑construction module. This module pairs each sentence with a structured prompt that includes definitions of the propaganda techniques described in~\cite{da2019fine}. The resulting prompt is then sent to an LLM (OpenAI; default model: \texttt{gpt-5.2-2025-12-11}\footnote{\url{https://platform.openai.com/docs/models}}), which returns both the predicted technique and a concise explanation of its reasoning. The model is configurable via environment variable, allowing operators to substitute any compatible OpenAI endpoint. The system supports three verbosity modes(\textit{low}, \textit{medium}, and \textit{verbose}), which control word limits on evidence and explanation fields, allowing users to trade response richness for latency and token cost. LLM outputs are validated against the closed 17-technique taxonomy; any out-of-taxonomy labels are silently dropped. API keys are provided by the user at runtime and are never stored persistently or transmitted beyond the OpenAI API endpoint.
The complete tool is packaged as a publicly available Docker container\footnote{\url{https://github.com/propasafe/propasafe-hybrid}}, making deployment straightforward. Fig.~\ref{fig:architecture_img} illustrates the overall architecture of Propasafe‑Hybrid. The top three components: Article Extractor, Offline Classification Module, and the fine‑tuned BERT model, are inherited from the original Propasafe system. Our contribution is the Explainability Module, which communicates with an LLM via API calls and enriches the offline predictions with technique‑level insights.

A key design decision concerns disagreement between the two stages. BERT's classification is authoritative: highlight color and severity are assigned from the BERT probability score before the LLM is ever invoked, and are never modified by the LLM response. When the LLM returns no recognized technique for a flagged sentence, either because it finds no propaganda or because its output falls outside the closed 17-technique taxonomy; the sentence remains highlighted with its BERT-derived color, and the \texttt{llm\char`_status} field is set to \texttt{empty}. This ensures visual feedback is never suppressed by LLM ambiguity, while the technique panel simply remains blank for that sentence. The \texttt{llm\char`_status} field distinguishes four states: \texttt{not\char`_run} (below threshold), \texttt{ok} (LLM confirmed $\geq$1 technique), \texttt{empty} (LLM returned no techniques), and \texttt{error} (API failure).

The hybrid workflow produces a multi‑layered output. The pre-filtering stage provides a measurable reduction in LLM token usage. Based on the system's behavior on representative news articles, approximately 80--87\% of sentences are filtered out by BERT before reaching the LLM. On a typical mid-length article of roughly 40 sentences, this means only 5--8 sentences are forwarded for LLM processing. The threshold is set at 0.50, the natural decision boundary, which maximizes recall at the cost of occasionally forwarding borderline sentences; these cases surface as \texttt{llm\char`_status=empty} in the output.

At the interface level, the system highlights sentences based on their offline BERT classification probability: values between 0.50 and 0.75 are highlighted in yellow (mild likelihood), while values above 0.75 are highlighted in red (high likelihood). These highlighted sentences are then annotated with the LLM‑identified propaganda technique and an explanation that justifies the classification.

\begin{figure}[!tbp]
    \centering
    \includegraphics[width=0.5\columnwidth]{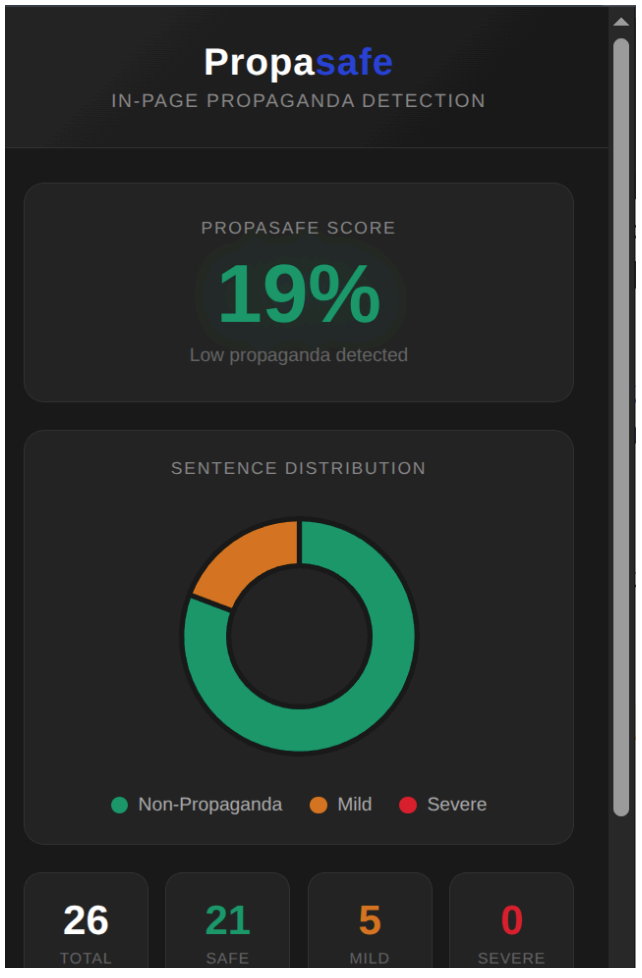}
    \alt{Dashboard showing Propasafe score of 19 percent with low propaganda detected. A donut chart displays sentence distribution with 26 non-propaganda, 21 mild, 5 severe, and 0 additional sentences.}
    \caption{Glance screen of Propasafe-Hybrid displaying the proportion of flagged sentences and a breakdown by severity level (safe, mild, severe).}
    \label{fig:glance_screen}
\end{figure}

Fig.~\ref{fig:glance_screen} displays the ``glance" view of the interface, showing the percentage of sentences flagged as propaganda, the total number of sentences in the article, and a breakdown of safe (probability $< 0.50$), mild ($0.50$--$0.75$), and severe ($> 0.75$) sentences.

\begin{figure}[!tbp]
    \centering
    \includegraphics[width=0.9\columnwidth]{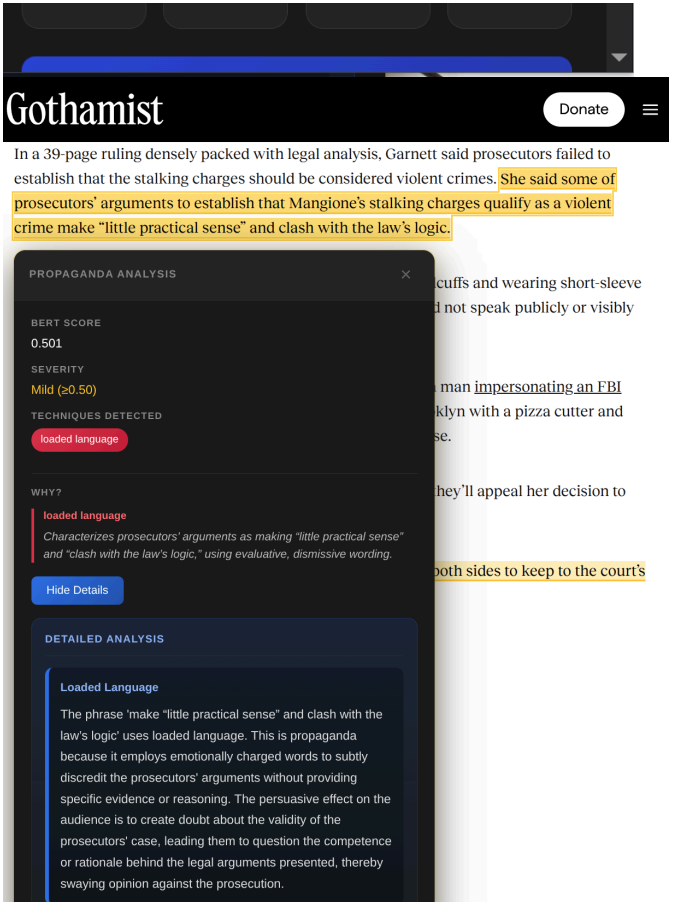}
    \alt{Screenshot of a news article with a sentence highlighted in yellow indicating mild propaganda. A sidebar panel shows BERT score 0.501, severity mild, technique detected as Loaded Language, with a why explanation and detailed analysis provided by the OpenAI API.}  
    \caption{Propasafe-Hybrid highlighting a mild propaganda instance (Loaded Language) and providing a technique-level explanation via the OpenAI API.}
    \label{fig:analyze1}
\end{figure}

Fig.~\ref{fig:analyze1} provides an example where Propasafe‑Hybrid identifies a sentence as exhibiting Loaded Language, and the LLM offers a corresponding explanation for why this technique is present.

\section{Conclusion and Future Work}
\label{sec:conclusion}

\begin{figure}[!tbp]
    \centering
    \includegraphics[width=0.9\columnwidth]{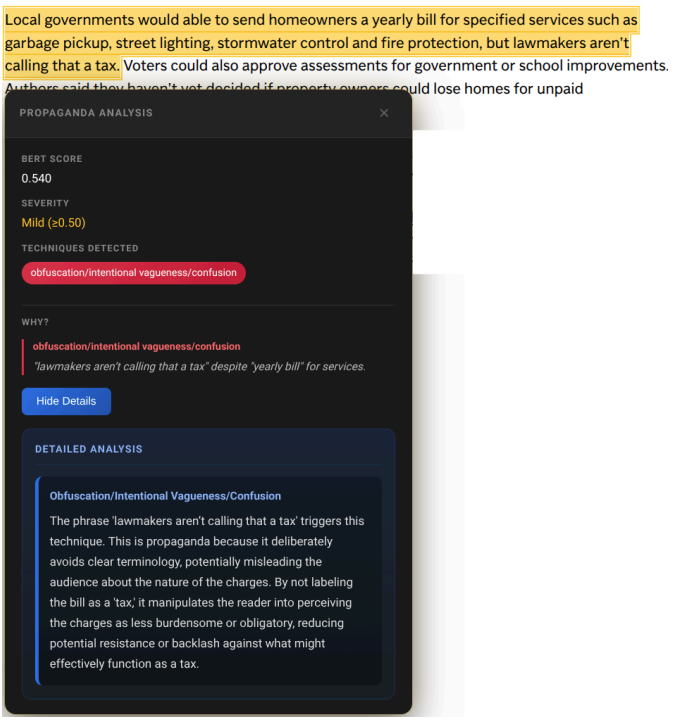}
    \alt{Screenshot of a news article with a sentence highlighted in yellow indicating mild propaganda. A sidebar panel shows BERT score 0.540, severity mild, technique detected as Obfuscation/Intentional Vagueness/Confusion, with a why explanation and detailed analysis provided by the OpenAI API.}
    \caption{Propasafe-Hybrid highlighting a mild propaganda instance (Obfuscation/Intentional Vagueness) and providing a technique-level explanation via the OpenAI API.}
    \label{fig:analyze2}
\end{figure}

Fig.~\ref{fig:analyze2} shows another mild instance where the LLM classifies a sentence under Obfuscation / Intentional Vagueness / Confusion, accompanied by the model's rationale for the classification.

We introduced Propasafe‑Hybrid, a practical system that fuses offline BERT filtering with selective online LLM explainability to deliver real‑time, span‑level insights where readers encounter content. By classifying sentences locally and sending only likely propagandistic spans for explanation, the tool lowers latency and API usage while preserving interpretability through technique labels and concise rationales. The interface surfaces color‑coded confidence cues and inline explanations, helping non‑expert readers understand not only what is propagandistic but also the explanation behind the classification.
Looking ahead, we will prioritize a Chrome Web Store release for click‑and‑use installation and conduct a user study comparing Propasafe‑Hybrid to representative alternatives on effectiveness, efficiency, and perceived transparency.

\section{Limitations}
\label{sec:limit}

At present, the tool supports inference through a single OpenAI model, which constrains cost and latency tuning across different workloads. This single‑provider dependency also makes the system vulnerable to model version changes and pricing fluctuations. As future work, we will add multi‑provider support. We also plan to let users store multiple API keys securely and choose a default model/key profile before each run. The current implementation defaults to a hybrid mode, which simplifies deployment but limits user control over privacy and reliability trade‑offs. Some use cases require fully offline processing, while others benefit from detailed explanations from online models. We plan to work on mode-selector(offline, hybrid, online) so users can pick the most appropriate setting for their context and network conditions. To strengthen verifiability, we will add a side‑by‑side comparison view that runs the same input through chosen models and modes and displays aligned outputs.

\section{Acknowledgement}
\label{sec:ack}
This work used computational resources provided by the National Artificial Intelligence Research Resource (NAIRR) Pilot under Allocation
NAIRR240383.

\bibliographystyle{flairs}
\bibliography{references}

@inproceedings{da2019fine,
  title={Fine-grained analysis of propaganda in news articles},
  author={Da San Martino, Giovanni and Yu, Seunghak and Barr{\'o}n-Cedeno, Alberto and Petrov, Rostislav and Nakov, Preslav},
  booktitle={Proceedings of the 2019 conference on empirical methods in natural language processing and the 9th international joint conference on natural language processing (EMNLP-IJCNLP)},
  pages={5636--5646},
  year={2019}
}

@inproceedings{da2020semeval,
  title={SemEval-2020 task 11: Detection of propaganda techniques in news articles},
  author={Da San Martino, Giovanni and Barr{\'o}n-Cede{\~n}o, Alberto and Wachsmuth, Henning and Petrov, Rostislav and Nakov, Preslav},
  booktitle={Proceedings of the fourteenth workshop on semantic evaluation},
  pages={1377--1414},
  year={2020}
}

@article{sharma2025analysis,
  title={Analysis of Propaganda in Tweets From Politically Biased Sources},
  author={Sharma, Vivek and Shokri, Mohammad Mahdi and Levitan, Sarah Ita and Filatova, Elena and Jain, Shweta},
  journal={arXiv preprint arXiv:2507.08169},
  year={2025}
}

@article{barron2019proppy,
  title={Proppy: Organizing the news based on their propagandistic content},
  author={Barr{\'o}n-Cedeno, Alberto and Jaradat, Israa and Da San Martino, Giovanni and Nakov, Preslav},
  journal={Information Processing \& Management},
  volume={56},
  number={5},
  pages={1849--1864},
  year={2019},
  publisher={Elsevier}
}

@inproceedings{sharma2025propasafe,
  title={Propasafe: A BERT-based offline tool for propaganda detection},
  author={Sharma, Vivek and Shokri, Mohammad Mahdi and Jain, Shweta and Levitan, Sarah Ita and Filatova, Elena},
  booktitle={Companion Proceedings of the ACM on Web Conference 2025},
  pages={2903--2906},
  year={2025}
}

@article{zavolokina2025biased,
  title={Biased by Design: Leveraging Inherent AI Biases to Enhance Critical Thinking of News Readers},
  author={Zavolokina, Liudmila and Sprenkamp, Kilian and Katashinskaya, Zoya and Jones, Daniel Gordon},
  journal={arXiv preprint arXiv:2504.14522},
  year={2025}
}

@inproceedings{zavolokina2024think,
  title={Think fast, think slow, think critical: designing an automated propaganda detection tool},
  author={Zavolokina, Liudmila and Sprenkamp, Kilian and Katashinskaya, Zoya and Jones, Daniel Gordon and Schwabe, Gerhard},
  booktitle={Proceedings of the 2024 CHI Conference on Human Factors in Computing Systems},
  pages={1--24},
  year={2024}
}

@article{martino2020survey,
  title={A survey on computational propaganda detection},
  author={Martino, Giovanni Da San and Cresci, Stefano and Barr{\'o}n-Cede{\~n}o, Alberto and Yu, Seunghak and Di Pietro, Roberto and Nakov, Preslav},
  journal={arXiv preprint arXiv:2007.08024},
  year={2020}
}

@inproceedings{da2020prta,
  title={Prta: A system to support the analysis of propaganda techniques in the news},
  author={Da San Martino, Giovanni and Shaar, Shaden and Zhang, Yifan and Yu, Seunghak and Barr{\'o}n-Cedeno, Alberto and Nakov, Preslav},
  booktitle={Proceedings of the 58th Annual Meeting of the Association for Computational Linguistics: System Demonstrations},
  pages={287--293},
  year={2020}
}

\end{document}